\documentclass{iopjournal}

\usepackage{amsmath}
\usepackage{amssymb}
\usepackage{booktabs}
\usepackage{tikz}
\usetikzlibrary{arrows.meta,positioning,calc,fit,backgrounds}
\usepackage[numbers,sort&compress]{natbib}
\usepackage{xspace}
\usepackage{siunitx}
\usepackage[inline]{enumitem}

\newcommand{\artist}{\texttt{artist}\xspace}

\begin{document}

\articletype{Technical Note}

\title{\artist: A Python package for differentiable ray tracing and gradient-based optimization of solar tower power plant operation}

\author{Marlene Busch$^{1,*}$\orcid{0009-0008-5730-7528},
Marie Weiel$^{2,3}$\orcid{0000-0001-9648-4385},
Kaleb Phipps$^{2,3}$\orcid{0000-0002-9197-1739},
Daniel Maldonado Quinto$^{1}$\orcid{0000-0003-2929-8667},
Max Pargmann$^{1}$\orcid{0000-0002-4705-6285}},
Robert Pitz-Paal$^{1,4}$\orcid{0000-0002-3542-3391}, and
Markus G\"otz$^{2,3}$\orcid{0000-0002-2233-1041}

\affil{$^1$Institute of Solar Research, German Aerospace Center (DLR), Cologne, Germany}

\affil{$^2$Scientific Computing Center (SCC), Karlsruhe Institute of Technology (KIT), Eggenstein-Leopoldshafen, Germany}

\affil{$^3$Helmholtz AI, Karlsruhe, Germany}

\affil{$^4$Chair of Solar Technology, RWTH Aachen University, Aachen, Germany}

\affil{$^*$Author to whom any correspondence should be addressed.}

\email{marlene.busch@dlr.de}


\begin{abstract}
Solar tower power plants provide dispatchable, renewable energy by concentrating sunlight with arrays of several thousand mirrors, called heliostats, onto a central receiver. 
Efficient operation requires accurate modeling of the optical path from each heliostat to the receiver, along with continuous compensation for heliostat manufacturing imperfections, kinematic misalignment, and material degradation that distort the concentrated flux distribution. 
Conventional ray tracers simulate this optical behavior by predicting a flux density distribution from an assumed plant model.
However, they are limited to forward simulation and cannot solve the inverse problem of recovering unknown plant parameters from measured flux distributions nor directly optimize plant operation.
We present \artist, an open-source, FAIR4RS-compliant Python package implementing a fully differentiable digital twin of solar tower power plants. 
By making the optical pipeline, comprising heliostat surface representation, heliostat kinematics, heliostat blocking, and ray tracing, end-to-end differentiable, \artist turns the digital twin from a passive forward model into an active physics-based tool for inverse modeling and data-driven optimization. 
The resulting gradient-based optimization framework handles tasks conventionally addressed by separate, specialized solutions: 
reconstructing heliostat surfaces and kinematics from operational flux images without costly deflectometry measurements; predicting irradiance at the receiver; and optimizing aim points across the heliostat field to realize a desired receiver flux density distribution. 
Built on PyTorch's automatic differentiation engine, \artist scales to real plants through native GPU acceleration, a structure-of-arrays memory layout, and nested data parallelism for distributed execution.
It interfaces with open data sources including the \emph{PAINT} database for operational plant data. 
We describe the software architecture, the differentiable physical models, and the unified optimization framework, and demonstrate the workflow on representative reconstruction and optimization tasks. 
By integrating modeling, prediction, and optimization into a single interpretable, openly available simulation package, \artist provides a principled foundation for data-driven research toward autonomous solar tower operation.
\end{abstract}

\section{Introduction}
Transitioning to a sustainable energy system built around renewables is essential to international strategies for climate change mitigation~\citep{edenhofer2011renewable}. 
Concentrating solar power (CSP) is a promising complement to photovoltaics for achieving this goal because, unlike photovoltaics, it can store thermal energy at relatively low cost~\citep{zhu2022roadmap, zhu2023heliocon}. 
Solar tower power plants are a particularly scalable form of CSP~\citep{alexopoulos2022concentrating}. 
A field of several thousand mirrors, known as heliostats, tracks the sun and reflects sunlight onto a receiver located on a central tower. 
The receiver heats a working medium, and the resulting thermal energy is either converted into carbon-neutral electricity or used in industrial processes~\citep{zhu2023heliocon}. 

Despite this potential, broad deployment is hindered by high investment and operating costs as well as demanding control requirements~\citep{huang2021survey,zhu2023heliocon}.
A key difficulty is that small imperfections in heliostats across the field accumulate and significantly affect the plant's optical performance.
Heliostat surface deformations caused by weathering and aging, together with kinematic misalignments arising from manufacturing, installation, and calibration errors, distort the flux density distribution on the receiver~\citep{ulmer2011automated,sattler2020review}.
This distortion is not negligible: a \SI{1}{\milli\radian} tracking error offsets the reflected beam by approximately one meter at a one‑kilometer distance, and accumulated optical errors can reduce energy yield by ten to twenty percent~\citep{zhu2023heliocon,yuan2015compensation}.
The many heliostats in a field make it impractical to measure each component manually and monitor how their deviations evolve over time.
Because these deviations cannot be estimated precisely, solar tower power plants rely on conservative safety margins and human oversight, limiting achievable operating temperatures and the degree of automation.
The large number of heliostats and their unknown component-level deviations turn efficient operation into a challenging, high-dimensional control problem, where each heliostat must be assigned an aim point such that the resulting flux density distribution on the receiver is optimal subject to thermal safety limits~\citep{belhomme2014optimization,oberkirsch2023validation}.

Digital twins of CSP plants constitute a powerful tool for meeting these operational requirements and enabling autonomous, cost-effective operation~\citep{rasheed2020digital,huang2021survey}. 
At the core of such a digital twin is a Monte Carlo ray tracer:
representative light rays propagate through a three-dimensional scene, reflect off heliostats, and concentrate on the receiver, where their energy is accumulated to obtain a flux density distribution.
Established CSP ray tracers~\citep{ahlbrink2012stral,wendelin2018soltrace,cardoso2018new,mitchell2025modeling} accurately predict the flux density distribution from a given plant model. 
From a machine-learning perspective, they are limited to \emph{forward} simulation: 
they map plant parameters to the resulting flux density distribution.
Conversely, \emph{inverse} rendering seeks to recover underlying plant parameters, such as heliostat surface geometry and kinematics, from measured flux images.
Because conventional ray tracers are not differentiable, they do not provide the gradients required for scalable inverse rendering or efficient gradient-based optimization of operational variables such as heliostat aim points.
This limits their applicability for plant operation because acquiring the physical parameters required for accurate prediction is too costly and time-consuming at plant scale.

 \begin{figure}[!b]
 \centering
 \includegraphics[width=\textwidth]{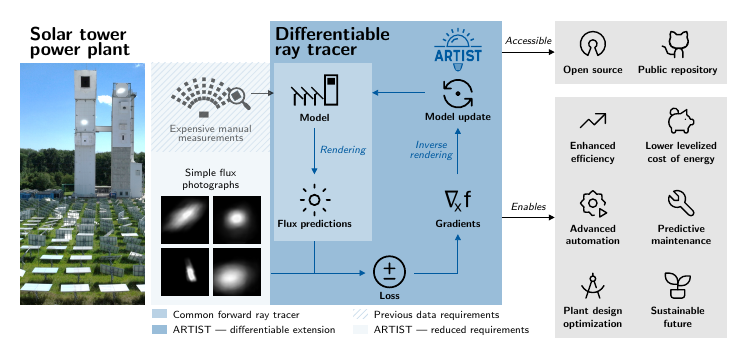}
 \caption{
 \textbf{Overview of ray tracing in solar tower power plant digital twins.} 
  Conventional ray tracers predict flux density distributions from a plant model (forward rendering). 
  \artist extends this workflow with differentiable inverse rendering, enabling recovery of unknown plant parameters from flux images acquired during normal operation, and direct gradient-based optimization of plant operation. 
  Together, these capabilities reduce reliance on expensive manual measurements, enabling more autonomous and efficient solar tower operation through next-generation control strategies.}
 \label{fig:flowchart}
\end{figure}

Differentiable simulation~\citep{newbury2024review} overcomes this limitation:
by formulating each step of the optical pipeline as a differentiable operation, gradients of any objective with respect to physical parameters become available, transforming a forward-only simulation model into a bidirectional framework for inverse modeling and gradient-based optimization.
Individual components of this approach have already been demonstrated for solar tower power plants~\citep{pargmann2024automatic,lewen2025inverse,kuhl2024flux,sattler2024development} but their differing formulations prevent straightforward integration into a common simulation framework.

We present \artist\footnote{\emph{AI-Enhanced Differentiable Ray Tracer for Irradiation Prediction in Solar Tower Digital Twins}}, an open-source, FAIR4RS-compliant\footnote{Short for Findable, Accessible, Interoperable, and Reproducible principles for research software.} Python package implementing a fully differentiable digital twin of solar tower power plants.
Built on PyTorch's automatic differentiation engine~\citep{ansel2024pytorch}, \artist enables end-to-end training of a physics-based simulation model covering the complete optical pipeline, including heliostat surface representation, kinematics, ray tracing, and blocking.
Figure~\ref{fig:flowchart} illustrates how this differentiable formulation extends the conventional forward ray tracing workflow with inverse modeling and optimization capabilities, creating a closed feedback loop for continuous refinement of the plant model from operational flux images.
\artist thereby reduces the reliance on predefined plant models and expensive manual measurements conventionally required to obtain accurate forward predictions.
The resulting differentiable framework enables the same gradient-based optimization approach to address various tasks for efficient solar tower operation.
In this work, we focus on three core applications:
\begin{enumerate*}[label=\roman*)]
    \item reconstructing heliostat surfaces and kinematics from operational flux images;
    \item predicting irradiance at the receiver; and
    \item concurrently optimizing heliostat aim points to achieve a desired flux density distribution.
\end{enumerate*}
Unlike black-box surrogates, \artist is based on explicit physics and therefore remains interpretable. 
Predictions can be traced back to physical parameters, supporting trust in machine-generated operational decisions.
Its open-source implementation and public repository make \artist widely accessible, while native GPU acceleration, distributed execution, and built-in interfaces to open data sources such as the \emph{PAINT} database for operational CSP plant data~\citep{phipps2025paint,phipps2026paint} support its application to real-world plants. 
Together, these capabilities provide a foundation for more efficient, autonomous, and data-driven solar tower operation.

Our contributions are: 
\begin{enumerate}[label=\roman*.]
  % SYSTEM LEVEL
  \item \textbf{An end-to-end differentiable digital twin for solar tower power plants.} 
  We unify heliostat modeling, forward simulation, and inverse problem solving within a single physics-based framework.
  % COMPONENT LEVEL
  \item \textbf{Differentiable formulation of the optical pipeline.}
  We use spline‑based heliostat surface representation, rigid‑body kinematics, ray tracing via soft binning, and blocking via linear bounding volume hierarchies.
  This enables gradient-based inverse modeling and control across the full optical chain.
  % ENGINEERING
  \item \textbf{A scalable open‑source implementation.} 
  \artist is GPU-accelerated, distributed, and released as a FAIR-compliant Python package\footnote{\url{https://github.com/ARTIST-Association/ARTIST}} with interfaces to operational CSP datasets.
\end{enumerate}
The paper is organized as follows:
\textbf{Section~\ref{sec:relatedwork}} reviews the current state of the art.
\textbf{Section~\ref{sec:design}} details the software architecture. 
\textbf{Section~\ref{sec:methods}} describes the differentiable physical models and optimization framework. 
\textbf{Section~\ref{sec:applications}} demonstrates the workflow on representative reconstruction and optimization tasks.  
\textbf{Sections~\ref{sec:discussion} and~\ref{sec:conclusion}} discuss the implications, limitations, and future directions of our work.

\subsection{Related work}
\label{sec:relatedwork}

Commercial and scientific CSP ray tracers, such as \emph{STRAL}~\citep{ahlbrink2012stral}, \emph{SolTrace}~\citep{wendelin2018soltrace}, \emph{Tonatiuh}~\citep{cardoso2018new}, and \emph{TieSOL}~\citep{mitchell2025modeling}, are mature, well validated, and show reasonable agreement in forward flux predictions across different tools~\citep{mitchell2025modeling}. 
However, they are limited to forward simulation and cannot reconstruct unknown heliostat parameters from observations. 
That is why they either assume idealized heliostats or rely on dedicated, and often impractical, metrology~\citep{arqueros2003novel, xu2020brief, burke2023deflectometry}. 
In addition, many of the most capable tools are proprietary, limiting reproducibility and community-driven development. 

Alongside classical simulators, machine-learning methods have been explored for individual tasks, including 
differentiable ray tracing for in-situ metrology~\citep{pargmann2024automatic}, 
inverse deep learning for heliostat surface prediction~\citep{lewen2025inverse}, 
StyleGAN-based flux prediction~\citep{kuhl2025accurate}, and neural-network-based kinematic calibration~\citep{sattler2024development}. 
For plant operation, aim point optimization has been tackled using metaheuristics such as ant colony optimization~\citep{belhomme2014optimization}, closed-loop control~\citep{oberkirsch2023validation}, and, more recently, reinforcement learning and neural network surrogates~\citep{carballo2025reinforcement}. 
Although promising, these approaches remain task-specific and rely on disparate methods that cannot be easily combined. 
Moreover, purely data-driven surrogates typically sacrifice interpretability and transfer poorly across plant configurations, requiring costly retraining.

\artist addresses these limitations through a unified digital twin built on a fully differentiable physical model.
It supports both forward prediction and inverse tasks, including reconstruction and optimization, while preserving the interpretability of classical ray tracers.

\section{Software design and architecture}
\label{sec:design}
\artist is a modular Python package designed for out-of-the-box use and straightforward extension with custom components.
A solar tower power plant is represented as a three-dimensional scene comprising a light source, a heliostat field, and one or more towers with receivers and calibration targets. 
The digital twin is constructed from a \emph{scenario} file that describes the plant's physical layout and state, including heliostat positions, kinematic configurations, receiver geometry, and, where available, measured surface and calibration data. 
\artist stores scenarios using the Hierarchical Data Format version 5 (HDF5)~\citep{hdf52024}, a self-describing hierarchical file format based on groups and datasets. 
Each scenario's root group contains separate groups for heliostats, solar tower target areas, light sources, and prototypes, providing a structured and extensible representation.
Scenario generators enable users to create scenario files from plant specifications, while dedicated parsers facilitate automatic construction from supported data sources. 
In particular, \artist provides a parser for the open-access database for operational CSP plant data, \emph{PAINT}~\citep{phipps2026paint,phipps2025paint}.
Given a scenario, \artist builds a fully differentiable digital twin that can be used for
\begin{enumerate}[label=\roman*.]
  \item \textbf{Reconstruction} of heliostat surfaces and kinematics from measured flux images,
  \item \textbf{Prediction} of the flux density distribution on the receiver, and
  \item \textbf{Optimization} of heliostat aim points and thus motor positions for the entire field.
\end{enumerate}
The resulting optimized parameters, in particular the motor positions for each heliostat, can be exported to plant control software.

\subsection{Usability through modularity and extensibility}

To accommodate diverse plant architectures, such as different heliostat types, kinematic mechanisms, and receiver designs, \artist employs a modular architecture based on abstract classes, inheritance, and composition.
Core components of the optical pipeline, including heliostat surfaces, kinematics, actuators, receivers, calibration targets, and the light source, are represented by abstract component classes.
Users can extend these interfaces through inheritance to introduce plant-specific models, while composition enables flexible assembly of complete plant configurations without modifying the underlying simulation pipeline.
This modular design facilitates adaptation to different plant layouts, simplifies the integration of new component models, and promotes reuse of individual simulation components across different applications.

\subsection{Inverse modeling and optimization through differentiable simulation}

\artist's key feature is its fully differentiable optical pipeline, which combines models of heliostat surfaces, kinematics, ray tracing, and blocking within a physics-based optimization framework.
Physical components are implemented as \texttt{torch.nn.Module} subclasses and expressed entirely through differentiable tensor operations, enabling the complete simulation pipeline to be represented as a computational graph within PyTorch~\citep{ansel2024pytorch}.
\artist thus integrates into the PyTorch ecosystem of automatic differentiation, gradient-based optimization, GPU acceleration, and established machine-learning infrastructure.
Optimizable parameters are explicitly registered, allowing gradients of scalar objectives, such as losses comparing predicted flux images against measurements, to be computed automatically and propagated efficiently through the highly vectorized simulation pipeline.
By coupling automatic differentiation with physically interpretable models, \artist enables the same simulation framework to perform forward prediction, measurement-driven inverse modeling, and gradient-based optimization while preserving physical consistency.
This capability forms the technical foundation for the unified methodology described in Section~\ref{sec:methods}.

\subsection{Scalability through parallelization and distribution}

\artist targets real-world plants, where the digital twin must produce results quickly enough to support near‑real‑time control.
Computations across the heliostat field are vectorized by storing all heliostat data in a \emph{Structure‑of‑Arrays} (SoA) layout~\citep{homann2018soax}, in which each array holds one parameter type, such as the position, for all heliostats.
Combined with the single‑instruction, multiple‑thread execution model of GPUs, this contiguous memory layout enables coalesced memory access and highly parallel tensor operations.

To accommodate heterogeneous heliostat fields, \artist groups heliostats by shared configuration, allowing efficient batch processing while preserving support for distinct heliostat parameter sets and alignment models.
For larger deployments, these heliostat groups can be distributed across multiple compute nodes, enabling parallel processing both across heliostat groups and within individual groups. 
Built on PyTorch, \artist leverages its native distributed-data-parallel framework and automatically selects the appropriate communication backend for the available operating system and hardware. 
The package is compatible with the execution backends and operating systems supported by PyTorch, including CPU and GPU execution on Windows and Linux.

\subsection{Reproducibility through FAIR4RS-compliant open software and data}

To promote reproducible research, \artist follows the FAIR principles for research software (FAIR4RS)~\citep{barker2022introducing} through open-source distribution, persistent versioned releases, comprehensive documentation, machine-readable scenario formats, and standardized interfaces to open datasets.
The software is maintained with version control and continuous testing to ensure portability and reproducibility.
Comprehensive documentation is hosted on \emph{Read the Docs}\footnote{\url{https://artist.readthedocs.io/en/latest/?badge=latest}.}, covering installation instructions, tutorials, mathematical and algorithmic foundations, and a complete API reference.
Compatibility with the open \emph{PAINT} database~\citep{phipps2026paint,phipps2025paint}, together with the standardized scenario format and parser, provides a reproducible pathway from publicly available operational CSP data to digital twin experiments, enabling researchers without access to a physical plant to develop and benchmark new methods for CSP operation.

\section{Differentiable models and optimization framework}
\label{sec:methods}

At the core of \artist are differentiable models of heliostat surfaces, kinematics, ray tracing, and blocking, integrated into a physics-based optimization framework that leverages gradient-based methods.
This combination stabilizes otherwise underdetermined inverse problems, reduces the effective search space, and enables data-efficient optimization with relatively few samples.

\subsection{Heliostat surface representation through NURBS}
\label{sec:nurbs}

Heliostat surfaces deviate from their ideal geometry due to manufacturing imperfections, gravitational sag, mounting stresses, and weathering, which distort each heliostat's reflected flux density distribution.
Since heliostat mirrors are typically assembled from multiple reflective facets, the deformation of each facet must be represented individually.
To account for these deformations, \artist models each heliostat facet as a Non-Uniform Rational B-Spline (NURBS) surface~\citep{piegl1997nurbs}. 
NURBS are piecewise polynomial, and hence differentiable, while providing a flexible representation of smooth surface deformations.
Each facet is represented by a weighted combination of control points $\mathbf{B}_{i,j}\in\mathbb{R}^3$, determined by NURBS basis functions and weights:
\begin{align}
\mathbf{S}(u,v) = \frac{\sum_{i=0}^{n}\sum_{j=0}^{m} N_{i,p}(u)\, M_{j,q}(v)\, w_{i,j}\, \mathbf{B}_{i,j}}
{\sum_{i=0}^{n}\sum_{j=0}^{m} N_{i,p}(u)\, M_{j,q}(v)\, w_{i,j}},
\qquad u\in[u_p,u_{n+p}],\; v\in[v_q,v_{m+q}]
\label{eq:nurbs}
\end{align}
$N_{i,p}(u)$ and $M_{j,q}(v)$ are the B-spline basis functions of degree $p$ and $q$, defined recursively over uniformly spaced knot vectors in \artist, and $w_{i,j}$ are scalar NURBS weights. 
Because the geometry is largely determined by the control points $\mathbf{B}_{i,j}$, they become the optimizable parameters learned from measured flux images during surface reconstruction. 
Surface normals required for ray tracing are computed analytically from the partial derivatives of the NURBS surface at sampled parametric coordinates $(u,v)$, preserving differentiability throughout the optical pipeline.
This GPU-accelerated implementation enables parallel evaluation of thousands of differentiable heliostat facets during both forward simulation and inverse optimization.

\subsection{Heliostat kinematics through rigid-body transformations}
\label{sec:kinematics}

A heliostat's orientation is determined by its actuation system, which defines the kinematic mapping from motor positions to heliostat pose.
Heliostats typically use multiple actuators to enable orientation changes in three-dimensional space.
Each actuator is described by fixed mechanical parameters, such as travel limits and steps-to-stroke conversion factors.
The kinematic mapping is modeled as a chain of rigid-body transformations, where the sequence of translations and rotations is determined by actuator configuration and mechanical design.

In an ideal heliostat, actuators are mounted perfectly, and the kinematic chain consists only of the translational offsets between actuator axes and the commanded rotations about these axes.
In practice, manufacturing and installation tolerances introduce small deviations between the nominal and actual actuator axes. 
While other components of the kinematic chain may also be uncertain, angular misalignments dominate the optical error and therefore provide a compact parameterization of kinematic deviations. 
To capture these effects, \artist augments the rigid-body transformations with additional rotation terms describing heliostat-specific actuator-axis tilts.

Since the rigid-body transformations are implemented as differentiable tensor operations, gradients flow through the kinematic chain back to the learnable axis-tilt parameters.
During kinematics reconstruction, these parameters are learned from measured calibration images via gradient-based optimization, yielding a physically consistent kinematic model.

\subsection{Differentiable ray tracing through soft binning}
\label{sec:raytracing}
\artist's forward optical model is a double-integral, importance-sampled ray tracer~\citep{pargmann2024automatic}. 
Rays are cast from a discrete grid on each heliostat's surface and reflected according to the local surface normals and the solar position. 
This heliostat-centric tracing scheme focuses computation on rays likely to hit the receiver and is widely used in CSP. 
While ray propagation itself is continuous, conventional ray tracing becomes non-differentiable during the final accumulation step.
Reflected rays are hard‑binned on the receiver, and each ray contributes its full energy to the nearest pixel of a discretized flux density map.
This assignment is a discontinuous function of the ray's landing position, yielding zero gradients almost everywhere and undefined gradients at pixel boundaries. 
\artist replaces hard binning with \emph{soft binning} based on bilinear splatting~\citep{zwicker2001surface}: 
each ray's energy is distributed linearly among its four nearest pixels according to its sub-pixel landing position. 
The resulting flux density distribution is a smooth, piecewise-linear function of the ray landing direction, letting gradients flow from a flux-based loss back through the receiver map and the surface normals to all upstream physical parameters. 
\artist thus preserves the efficiency of the double-integral scheme while rendering the entire optical pipeline differentiable.

\subsection{Differentiable ray blocking through soft occlusion}
\label{sec:blocking}
When the heliostat field is densely packed or the sun is low on the horizon, one heliostat can block rays reflected by another before they reach the receiver.
Accurately modeling this effect requires evaluating intersections between millions of rays and thousands of potential blocking heliostats.
To address this computational challenge, \artist implements a fully differentiable blocking model based on a Linear Bounding Volume Hierarchy (LBVH) acceleration structure~\citep{karras2012maximizing}.

Each heliostat is approximated as one rectangular blocking primitive. 
The primitives are encoded with Morton codes derived from their centroids, sorted, and assembled into a binary radix tree in a data-parallel fashion~\citep{karras2012maximizing}. 
Axis-aligned bounding boxes (AABBs) are then computed for all internal tree nodes.
During ray tracing, the LBVH is traversed using ray-AABB intersection tests to efficiently identify potential blocking heliostats.
For these candidates, \artist replaces the binary blocked-or-visible decision with a differentiable soft occlusion model.
For each ray and blocking primitive, the ray-plane intersection is computed. 
Sigmoid functions provide differentiable approximations of whether the intersection lies in front of the ray origin and within the blocking primitive.
These soft occlusion contributions are then accumulated using a Beer-Lambert transmittance model~\citep{pharr2023physically}.
The resulting occlusion signal is smooth and differentiable, allowing gradients to flow back to all upstream physical parameters.

\subsection{Unified optimization through task-specific objective functions}
\label{sec:optimization}

\artist's key advantage is that all reconstruction and optimization tasks are formulated within a common differentiable simulation framework.
Rather than designing a dedicated solver for each application, \artist recovers unknown physical parameters and optimizes controllable ones through gradient-based minimization with, e.g., Adam~\citep{kingma2015adam}.
For image-based tasks, measured and predicted flux density distributions, $\mathbf{F}_{\text{ref}}$ and $\mathbf{F}_{\text{pred}}$, are normalized and interpreted as discrete probability distributions.
Their agreement is measured using the Kullback-Leibler (KL) divergence:
\begin{align}
L_{\text{flux}} = D_{\text{KL}}\!\left(\mathbf{F}_{\text{ref}} \,\|\, \mathbf{F}_{\text{pred}}\right)
= \sum_{p} \mathbf{F}_{\text{ref},p}\,\log\frac{\mathbf{F}_{\text{ref},p}}{\mathbf{F}_{\text{pred},p}},
\label{eq:klflux}
\end{align}
where the summation runs over all receiver pixels $p$.
Task-specific regularizers and constraints can be added to the objective function as required.
These include, for example, smoothness priors for surface reconstruction, energy-conservation constraints, or operational constraints such as limiting spillage and maximum receiver flux.
Constraints are enforced using an augmented-Lagrangian formulation~\citep{bertsekas1999nonlinear}, and optimization variables are normalized to ensure stable optimization across different parameter scales.

This unified optimization strategy allows the same differentiable digital twin to solve diverse inverse problems by changing only the objective function, while the underlying forward model and optimization procedure remain unchanged.

\section{Example applications}
\label{sec:applications}

\artist's primary application is inverse modeling, where gradients are used to infer unknown plant properties from measured flux images and optimize operational parameters to achieve a desired flux density distribution. 
This enables data-driven decision support using the physics-based digital twin, providing actionable recommendations for plant operation. 

The measured flux images are acquired using the camera target method~\citep{baheti1980cameratarget}, in which individual heliostats are directed onto a Lambertian calibration target and imaged by a camera. 
Because the target reflects light diffusely, the recorded image intensity is proportional to the incident flux density distribution.

\begin{figure}[t]
    \centering
    \includegraphics[width=\textwidth]{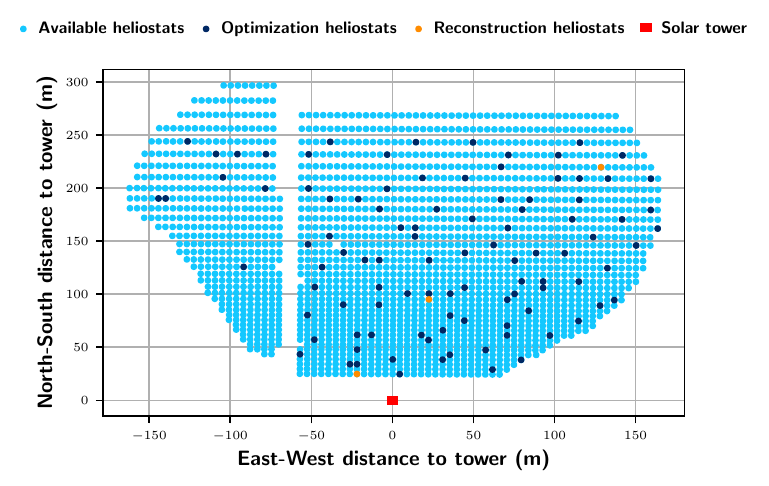}
    \caption{\textbf{Heliostat field layout of the Jülich solar tower power plant.}
    Heliostat positions are represented by light blue dots. 
    The heliostats used for the exemplary reconstruction are highlighted in orange, while those used for the exemplary aim point optimization are marked by dark blue dots.
    }
    \label{fig:field_layout}
\end{figure}

The following examples demonstrate the three main capabilities of \artist: reconstruction of unknown plant parameters, prediction of flux density distributions, and optimization of operational variables.
All examples are based on heliostats from the Jülich solar tower power plant in Germany\footnote{\url{https://www.dlr.de/en/research-and-transfer/research-infrastructure/solar-towers-juelich}} (see Figure~\ref{fig:field_layout}).
The results shown are representative demonstrations of \artist's capabilities rather than an exhaustive benchmark.

\subsection{Heliostat surface reconstruction from flux images}

By combining the NURBS surface representation (see Section~\ref{sec:nurbs}) with differentiable ray tracing (see Section~\ref{sec:raytracing}), \artist reconstructs heliostat surface deformations directly from measured flux images.
This reduces the need for dedicated deflectometry campaigns, drastically lowering the effort and cost associated with heliostat characterization and enabling more efficient and competitive power plant operation through improved simulation accuracy~\citep{ulmer2011automated,gonzalez2021calculating}.
For each heliostat, measured flux images are compared with the predicted flux density distribution using the KL divergence from Equation~\eqref{eq:klflux}.
The resulting gradients are used to update the NURBS control points, allowing surface deformations to be recovered starting from an ideal flat surface.

Differences between reference flux images, $\mathbf{F}_{\text{ref}}$, and predictions obtained from nominal (ideal) or reconstructed surfaces, $\mathbf{F}_{\text{pred}}$, are quantified using the normalized absolute flux error:
\begin{equation}
    E_{\text{L1}}
    =
    \frac{\sum_p
    \left|
    \mathbf{F}_{\text{pred},p}
    -
    \mathbf{F}_{\text{ref},p}
    \right|}
    {\sum_p\mathbf{F}_{\text{ref},p}},
\end{equation}
where the summation runs over all receiver pixels $p$ and flux images are normalized before comparison. 
The metric therefore evaluates only differences in the spatial flux distribution and is independent of the absolute flux magnitude.

Averaged over all evaluated heliostats and all flux images acquired for each heliostat, the normalized absolute flux error decreases from \(E_{\text{L1, ideal}} = \num{0.55}\) for the nominal surface representation to \(E_{\text{L1, reconstructed}} = \num{0.15}\) for the reconstructed NURBS surface. 
The normalized absolute flux error is reduced by \SI{72.7}{\percent}, or a factor of \num{3.6}. 
These results demonstrate that incorporating reconstructed surface models substantially improves the agreement between simulated and measured flux density distributions.
Note that the reported improvement should be interpreted as a conservative estimate of the achievable performance gain. 
The effect of surface deformations on the flux density distribution depends strongly on heliostat position and solar position.

\subsection{Heliostat kinematics reconstruction and calibration}
\label{sec:kinematics_reconstruction}
\artist calibrates heliostat kinematics by reconstructing the learnable actuator-axis tilts introduced in Section~\ref{sec:kinematics} directly from measured flux images using gradient-based optimization. 
This extends the conventional camera target method~\citep{baheti1980cameratarget, waghmare2022center, king1981heliostat}, which establishes an initial aiming accuracy for each heliostat, by estimating residual alignment errors from the measured flux images.

Two complementary reconstruction strategies are available:
\begin{enumerate}[label=\roman*.]
    \item \textbf{Alignment-driven reconstruction.} 
    Focal spot positions are extracted from the calibration images and converted into reference heliostat normals using the known incident ray direction, heliostat position, and law of reflection. 
    The kinematic model predicts the heliostat normal from the recorded actuator positions and current kinematic parameters. 
    The optimization minimizes the difference between the predicted and reference normals by iteratively updating the kinematic parameters until the observed heliostat alignment is reproduced.

    \item \textbf{Flux-driven reconstruction.} 
    Alternatively, the optimization can be performed directly on the measured flux images. 
    At each optimization step, the current kinematic parameters are used to align the heliostat, and the differentiable ray tracer predicts the resulting flux density distribution on the receiver. 
    The focal spot is extracted from both predicted and measured flux images, and their Euclidean distance is used as the optimization loss.
    Gradients are propagated through the ray tracer, NURBS surface model, and kinematic chain to update the kinematic parameters and reduce the misalignment between predicted and measured flux density distributions.
\end{enumerate}
Both approaches are initialized from the plant's measured configuration as a physically meaningful starting point. 
Alignment-driven reconstruction is computationally efficient because it avoids ray tracing and remains applicable for large initial calibration errors, where reflected rays may miss the receiver entirely and no flux-based loss can be evaluated.
However, converting the measured focal spot into a reference normal introduces a small geometric approximation, since the focal spot centroid only approximates the true aim point.
It therefore provides a fast and robust initial calibration.
Flux-driven reconstruction can subsequently refine the kinematic parameters by directly minimizing the discrepancy between predicted and measured flux images once the predicted flux reliably hits the receiver.

\subsection{Flux density distribution prediction}
\label{sec:flux_prediction}
The reconstructed heliostat surfaces and kinematics provide a more accurate simulation model for forward flux prediction than idealized heliostat models.
By combining the reconstructed physical parameters with GPU-accelerated ray tracing, \artist computes the superposition of flux density distributions from thousands of heliostats while maintaining the scalability required for full-field simulations.

\begin{figure}[t]
    \centering
    \includegraphics[width=\textwidth]{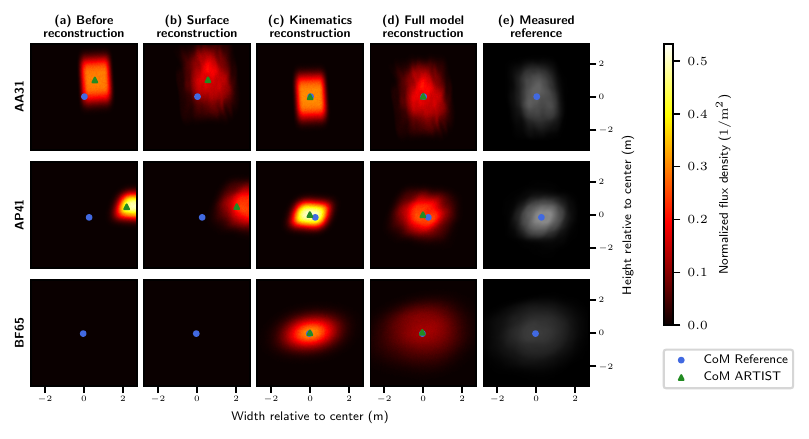}
    \caption{\textbf{Predicted and measured flux images for three heliostats: (a)} idealized model, \textbf{(b)} reconstructed surface only, \textbf{(c)} reconstructed kinematics only, \textbf{(d)} full reconstructed model combining reconstructed surface and kinematics, and \textbf{(e)} measured reference flux image. 
    The full reconstructed model shows the closest agreement with the measured flux density distribution.
    Within each row, the flux density distributions are normalized such that the integrated reflected power originating from the heliostat area equals one for the full reconstructed model.
    Flux density distributions that are partially off-target therefore contain only a fraction of the total flux and have an integral smaller than one.
    All images are shown on the same spatial scale, corresponding to the physical dimensions of the calibration target, with the image axes centered at the receiver center. 
    }
    \label{fig:model_reconstruction}
\end{figure}

Figure~\ref{fig:model_reconstruction} illustrates how different stages of model reconstruction affect the predicted flux density distributions for three heliostats of the Jülich solar tower power plant (marked by the orange dots in Figure~\ref{fig:field_layout}), compared with the measured reference in panel~\ref{fig:model_reconstruction}(e).
Panel~\ref{fig:model_reconstruction}(a) assumes perfect geometry and kinematics, yielding undistorted flux density distributions that are misaligned with the measured reference, with heliostat AP41 partially and heliostat BF65 completely missing the target. 
Panel~\ref{fig:model_reconstruction}(b) incorporates reconstructed surface parameters, capturing mirror deformations while leaving alignment unchanged. 
Conversely, panel~\ref{fig:model_reconstruction}(c) reconstructs the heliostat kinematics, correcting the center of mass (CoM) of the predicted flux density distribution while retaining an idealized mirror surface. 
Finally, panel~\ref{fig:model_reconstruction}(d) combines reconstructed surface and kinematics parameters, closely matching the measured reference in both CoM alignment and flux density distortions. 
The results show that including both reconstructed surface and kinematic parameters substantially improves the agreement between simulated and measured flux density distributions compared with idealized models.

\subsection{Aim point optimization}
\label{sec:aimpoint_optimization}

Accurate heliostat models are essential for plant operation, as surface deformations and kinematic errors directly affect the receiver flux density distribution that can be achieved.
Previous studies have shown that incorporating reconstructed heliostat models into plant operation can increase energy yield by up to \SI{20}{\percent}~\citep{pargmann2024automatic}.
While reconstruction of individual heliostats is important, plant performance ultimately depends on their combined flux density distribution on the receiver.
Aim point optimization aims to achieve a desired receiver flux distribution while satisfying operational constraints and temperature limits.
Reducing local flux peaks and avoiding excessive thermal gradients mitigates thermal stresses and reduces material degradation, thereby lowering maintenance requirements and increasing receiver longevity.
Aim point optimization therefore enables operation closer to receiver thermal limits, increasing the usable energy delivered to the receiver while maintaining a controlled heat load distribution within safe operating limits~\citep{sanchez2018aiming, gonzalez2021calculating}.

To realize coordinated aim points, \artist directly optimizes heliostat motor positions for a desired receiver flux density distribution, simultaneously accounting for heliostat kinematics and surface imperfections without assuming ideal reflections.
The objective function combines the KL divergence from Equation~\eqref{eq:klflux} between desired and predicted flux density distributions with augmented-Lagrangian constraints that enforce flux conservation, limit spillage, and prevent local heat peaks.
Unlike approaches that optimize heliostats independently or restrict aim points to discrete candidate locations, \artist performs a joint optimization over continuous aim point positions, enabling a coordinated aiming strategy.

\begin{figure}
    \centering
    \includegraphics[width=\textwidth]{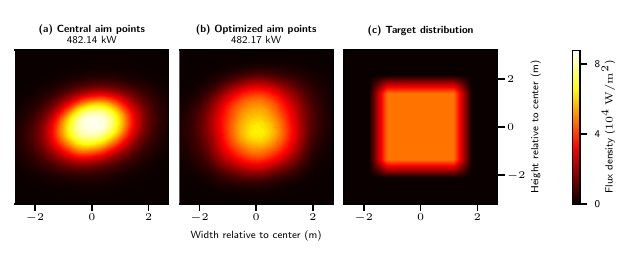}
    \caption{\textbf{Aim point optimization flux density distributions:} 
    \textbf{(a)} Flux density distribution with all heliostats aiming at the receiver center, 
    \textbf{(b)} optimized distribution obtained by assigning individual aim points to each heliostat, and 
    \textbf{(c)} the target distribution. 
    The optimization redistributes the incident flux density to approximate the target distribution while preserving the total incident power. 
    A total of \num{100} reconstructed heliostats were used for the optimization.
    All images are shown on the same spatial scale, corresponding to the physical dimensions of the target, with the image axes centered at the receiver center.
    }
    \label{fig:aim_point_optimization}
\end{figure}

Figure~\ref{fig:aim_point_optimization} shows an exemplary aim point optimization for \num{100} reconstructed heliostats of the Jülich solar tower power plant (marked by the dark blue dots in Figure~\ref{fig:field_layout}). 
Compared with the baseline case, in which all heliostats aim at the receiver center, the optimized aim points redistribute the incident flux density to closely match the target distribution while preserving a total incident power of approximately \SI{482}{\kilo\watt}.

\section{Discussion}
\label{sec:discussion}

Unifying prediction, reconstruction, and optimization within a fully differentiable, physics-based digital twin provides two main advantages.

First, the same forward model supports both simulation and inverse tasks.
The ray tracer used to predict flux images can be inverted to reconstruct heliostat surfaces and kinematics from measurements, and can also be used directly for operational optimization.
Compared with classical ray tracers~\citep{ahlbrink2012stral,wendelin2018soltrace,cardoso2018new,mitchell2025modeling}, \artist preserves the accuracy and interpretability of physics-based simulation while enabling inversion and optimization within the same framework.
This removes the need for separate tools and ensures consistency across tasks.

Second, the explicit representation of plant components keeps the resulting decisions interpretable.
Predictions can be traced back to meaningful model parameters, improving the transparency of machine-generated operational recommendations.
Compared with black-box machine-learning approaches~\citep{lewen2025inverse,kuhl2024flux}, \artist requires fewer training data and can be adapted to new plant configurations by updating model parameters rather than retraining surrogate models.
The combination of gradient-based optimization and physical constraints makes otherwise underdetermined inverse problems tractable.
These constraints reduce the search space and regularize the solution, enabling meaningful reconstruction from the limited flux measurements available during routine operation.

Several limitations define directions for future work.
The current optical model does not include shading, where incoming rays are blocked by neighboring heliostats before reflection. 
Since the implemented soft occlusion model is independent of whether rays are incoming or reflected, it provides a direct basis for extending \artist to differentiable shading.
\artist also lacks uncertainty quantification, which is important for reliable decision-making in high-temperature environments and for safe deployment of optimized motor positions in operational plants.
Although individual components build on previously validated approaches~\citep{pargmann2024automatic}, further validation across additional operational datasets and live plant measurements is required.
Finally, the differentiable formulation opens opportunities for inverse plant design, such as optimizing heliostat field layouts and receiver geometries for a desired output, which is beyond the scope of this work.

\section{Conclusion}
\label{sec:conclusion}

We have presented \artist, an open-source Python package implementing a fully differentiable digital twin of solar tower power plants. 
By making the optical pipeline differentiable, \artist extends the digital twin from a forward simulator to a tool for gradient-based inverse modeling and optimization. 
A single differentiable, physics-based framework enables surface and kinematics reconstruction from operational flux images, high-fidelity prediction of flux density distributions, and multi-heliostat aim point optimization, tasks traditionally addressed with separate tools and algorithms.
\artist is designed for the scale of real plants through GPU acceleration, a structure-of-arrays memory layout, and distributed computation. 
It follows the FAIR4RS principles and interfaces with the open \emph{PAINT} database, enabling reproducible workflows based on publicly available data.
By integrating prediction, reconstruction, and optimization in a single interpretable and openly available simulation framework, \artist provides a foundation for data-driven and physics-based research toward more autonomous and cost-effective solar tower operation. 
Future work will integrate differentiable shading and uncertainty quantification, broaden validation using more diverse operational data, and extend the framework toward inverse plant design.

\ack{
The authors thank the wider ARTIST development community for discussions and feedback. 
Computing resources were provided by HAICORE@KIT.
}

\funding{
This work is supported by the Helmholtz Association Initiative and Networking Fund through the Helmholtz AI platform and the ARTIST project under grant number ZT-I-PF-5-159.
}

\roles{Using the CRediT taxonomy: 
\textbf{Marlene Busch}: conceptualization, methodology, software, investigation, writing---original draft. 
\textbf{Marie Weiel}: methodology, software, writing---original draft. 
\textbf{Daniel Maldonado Quinto}: methodology, software, writing---review and editing. 
\textbf{Kaleb Phipps}: methodology, software, data curation, writing---review and editing. 
\textbf{Robert Pitz-Paal}: supervision, resources, writing---review and editing. 
\textbf{Markus G\"otz}: supervision, resources, methodology, writing---review and editing. 
\textbf{Max Pargmann}: conceptualization, methodology, supervision, writing---review and editing. All authors read and approved the manuscript.}

\data{\artist is released as open-source software under the MIT license and is available, together with documentation, from its public repository. The package interfaces with the open-access \emph{PAINT} database~\citep{phipps2026paint,phipps2025paint}, on which the described workflow can be reproduced.}

\section*{AI usage disclosure}
Generative AI was used as an editorial and technical aid, specifically for refining the manuscript text and for debugging code. 
It did not contribute to the architectural design of the software or to any scientific findings.

\section*{Conflict of interest}
The authors declare no conflict of interest.

\bibliographystyle{unsrtnat}
\bibliography{references}

\end{document}